\documentclass[conference]{IEEEtran}
\IEEEoverridecommandlockouts
\usepackage{bm}
\usepackage{cite}
\usepackage{amsmath,amssymb,amsfonts}
\usepackage{algorithmic}
\usepackage{graphicx}
\usepackage{textcomp}
\usepackage{xcolor}
\usepackage{multirow}
\usepackage{relsize}
\usepackage{algorithm}
\usepackage{xspace}
\usepackage{color,soul}
\usepackage{comment}
\usepackage{subcaption}
\usepackage{wrapfig}
\usepackage{dsfont}
\usepackage{flushend}

\let\origNabla\nabla
\renewcommand*\nabla{\mathlarger\origNabla}

\newcommand*{\balancecolsandclearpage}{%
  \close@column@grid
  \cleardoublepage
  \twocolumngrid
}

\def\BibTeX{{\rm B\kern-.05em{\sc i\kern-.025em b}\kern-.08em
    T\kern-.1667em\lower.7ex\hbox{E}\kern-.125emX}}
\begin{document}

\title{Multi-task Learning for Source Attribution and Field Reconstruction for Methane Monitoring\\
}
\author{\IEEEauthorblockN{Arka Daw \textsuperscript{\textsection}} 
\IEEEauthorblockA{\textit{Dept. of Computer Science} \\
\textit{Virginia Tech}\\
Blacksburg, United States \\
darka@vt.edu}
\and
\IEEEauthorblockN{Kyongmin Yeo}
\IEEEauthorblockA{\textit{IBM T.J. Watson Research Center} \\
Yorktown Heights, United States \\
kyeo@us.ibm.com}
\and
\IEEEauthorblockN{Anuj Karpatne}
\IEEEauthorblockA{\textit{Dept. of Computer Science} \\
\textit{Virginia Tech}\\
Blacksburg, United States \\
karpatne@vt.edu}
\and
\IEEEauthorblockN{Levente Klein}
\IEEEauthorblockA{\textit{IBM T.J. Watson Research Center} \\
Yorktown Heights, United States \\
kleinl@us.ibm.com}
}
\maketitle
\begingroup\renewcommand\thefootnote{\textsection}
\footnotetext{Work done during internship at IBM T.J.Watson Research Center.}
\endgroup

\begin{abstract}
  Inferring the source information of greenhouse gases, such as methane, from spatially sparse sensor observations is an essential element in mitigating climate change. While it is well understood that the complex behavior of the atmospheric dispersion of such pollutants is governed by the Advection-Diffusion equation, it is difficult to directly apply the governing equations to identify the source location and magnitude (inverse problem) because of the spatially sparse and noisy observations, i.e., the pollution concentration is known only at the sensor locations and sensors sensitivity is limited. 
  Here, we develop a multi-task learning framework that can provide high-fidelity reconstruction of the concentration field and identify emission characteristics of the pollution sources such as their location, emission strength, etc. from sparse sensor observations. 
  We demonstrate that 
  our proposed framework is able to achieve accurate reconstruction of the methane concentrations from sparse sensor measurements as well as precisely pin-point the location and emission strength of these pollution sources.
\end{abstract}


\section{Introduction}
Methane is one of the potent greenhouse gasses \cite{howarth2011methane, howarth2014bridge, schulze2009importance} that is emitted into the atmosphere through leakages in natural gas systems, raising livestocks, or via natural sources such as wetlands.  These methane emissions caused by human activities have further been identified as a major contributor to climate change \cite{cao1998global, reay2010methane, o2010possible}. The global warming potential of methane is 25 times larger than CO$_2$, over a 100 year, and most emission are preventable with monitoring and sensing solutions \cite{boucher2009indirect}.Thus, inferring the location and emission strength of pollution sources such as methane leaks are both essential in monitoring the air-quality as well as mitigating the climate change. However, one of the major challenges in localization of these emission sources is the unavailability of high-resolution methane concentration maps, and these individual pollution sources are to be estimated from the limited concentration measurements from a sparse sensor network. 

Digital twins  are information system that create digital replication of the physical  state and temporal evolution constrained by observations and the governing laws of physics\cite{niederer2021}. Planetary digital twins are proposed to be developed to better understand the long term impact of climate change \cite{bauer2021digital}. Digital twins can model the complexity of the forward propagation of multi source dispersion \cite{karion2019intercomparison} and enable a digital playground to study the best measurement approach to reconstruct information and identify sources. The digital twin can be informed by sensors and satellite measurements as realistic constrain of methane dispersion\cite{van2020satellite}. Here we demonstrate how a dynamic, two dimensional propagation of pollution plumes \cite{wang2021s3rp, wang2020sr}, under realistic scenario can be used to reconstruct the propagation field and identify sources using a data driven approach.

Here, we consider atmospheric inverse models that aim to either reconstruct the concentration field \cite{Liu2022} or identify pollution source information \cite{hwang2019bayesian} of airborne pollutants. While atmospheric inverse models have been studied extensively, due to the ill-posed nature of the inverse problem, it still remains as a challenging topic. Recently, potential of deep learning approaches for the inverse problem has been demonstrated \cite{fukami2021global} \cite{milletarisource}. We are interested in developing a deep-learning-based atmospheric inverse model, utilizing a digital twin to generate a range of possible emission scenarios.


We develop a multi-task learning framework that can provide a high-fidelity reconstruction of the spatio-temporal concentration field and identify the emission characteristics of the pollution sources such as their location, and emission strengths from a time series of sensor measurements. We propose a novel 3D diffusive masked convolution to gradually propagate the information from the sensor locations to the unobserved field using a diffusion process. We demonstrate that our multi-task model is able to achieve accurate reconstruction of the pollution concentrations from sparse sensor measurements as well as precisely pin-point the location and emission strength of these pollution sources.


\begin{figure*}[ht]
    \centering
    \includegraphics[width=1.0\textwidth]{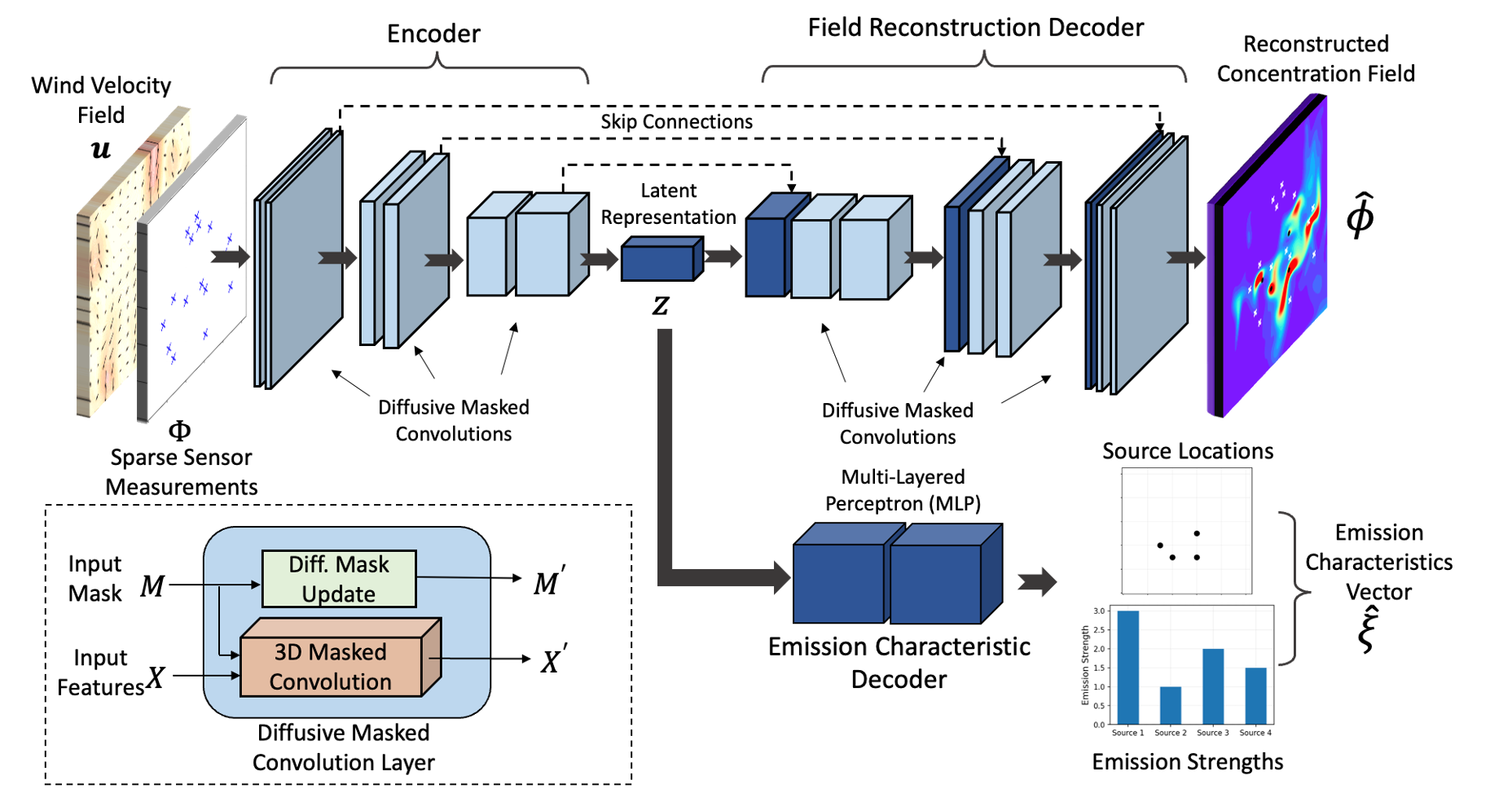}
    \caption{Neural Network Architecture of our proposed multi-task learning framework.}
    \label{fig:nn_arch}
\end{figure*}

\section{Background and Problem Setup}
Here, we assume that there is a digital twin for the domain of interest, $\Omega$. The digital twin consists of a numerical weather simulation model, that can generate a meteorological condition, such as the wind velocity field, $\bm{u}$, a computational model for atmospheric dispersion, and prior information, such as the locations of the potential emission sources, possible emission scenarios, and the locations of the sensors. We assume a two-dimensional domain, which is discretized by equispaced $N_x \times N_y$ grid points in the east-west ($x$) and the north-south ($y$) directions.

In the digital twin, the concentration for a particular emission scenario can be computed by solving the following advection-diffusion equation,
\begin{align}
\label{eq:adv_diff_eqn}
    \frac{\partial \phi}{\partial t}  + \mathbf{u\cdot} \nabla \phi - K \nabla^2 \phi = \sum_{i=1}^{N_s} q_i(\mathbf{x}, t); \:\: \mathbf{x} \in \Omega, t \in \mathbb{R}^+
\end{align}
where $\phi$ is the concentration, $\mathbf{u}$ is the wind velocity field, $K$ is the turbulent diffusivity, $q_i(\bm{x}, t)$ is the emission strength of the $i-th$ source, and $N_s$ denote the total number of sources. The emission sources are point sources; $q_{i}(x, t) = c_i\delta(\bm{x}-\bm{x}_i^s)$, where $c_i$ is the emission strength, $\bm{x}^s_i$ is the location of the source, and $\delta(.)$ is the Dirac delta function. 


Let the time-series of measurements from the $N_o$ sensors be $\bm{\Phi} = \{\bm{\Phi}_1, ..., \bm{\Phi}_{N_o}\}$, where $\Phi_{i}$ denotes the time series of the measurements of the $i$-th sensor, i.e., $\bm{\Phi}_{i}=[\Phi_{i}^1, \Phi_{i}^2, ..., \Phi_{i}^{N_t}]$, with $N_t$ denoting the number of time steps. The location of the $i$-th sensors are denoted by $\mathbf{x}_i^{\Phi}$. 

\emph{Problem Statement}: Given the time series of the sensor measurements $\Phi \in \mathbb{R}^{N_o \times N_t}$, their locations $\mathbf{x}^{\Phi} = \{\mathbf{x}^{\Phi}_1, \mathbf{x}^{\Phi}_2, ..., \mathbf{x}^{\Phi}_{N_o}\}$, and the wind velocity field $\mathbf{u}$, we aim to estimate the concentration $\phi$ on the $N_t \times N_x\times N_y$ spatio-temporal grid.
Additionally, we are also interested in estimating the emission characteristics of the $N_s$ potential emission sources, such as their constant emission strengths $\mathbf{c}=[c_1, c_2, ..., c_{N_s}]$, and their locations $\mathbf{x^s}=\{\mathbf{x}^s_1, \mathbf{x}^s_2,..., \mathbf{x}^s_{N_s}\}$. We assume that the digital twin platform can be used to generate the data on the $N_t\times N_x\times N_y$ grid for all possible emission scenarios.

\vspace{1ex}
\section{Proposed Method}
\label{sec:methods}

\subsection{Multi-task learning framework}
Multi-task learning \cite{zhang2021survey, standley2020tasks} is a learning paradigm, where the knowledge from one task can be utilized to improve the performance of the model on other similar tasks. Here, we aim to leverage the strong correlation between the two tasks of identifying the source information and reconstructing the concentration field. Note that, through the advection-diffusion equation, the concentration field is strongly coupled with the source characteristics given the wind velocity field $\mathbf{u}$.
We propose to learn a shared encoder $g(\boldsymbol{\theta}): [\Phi, \mathbf{x}^{\Phi}, \mathbf{u}] \rightarrow \mathbf{z}$ which aims to learn a latent representation $\mathbf{z}$. Then, this latent representation $\mathbf{z}$, is fed into task-specific decoders $f_1(\boldsymbol{\omega}): \mathbf{z} \rightarrow \phi$ and $f_2(\boldsymbol{\psi}): \mathbf{z} \rightarrow [\mathbf{c}, \mathbf{x}^s]$ to find the solution of the inverse problem and obtain the reconstructed field.

The multi-task learning framework can then be optimized using the objective function $\mathcal{L}(\boldsymbol{\theta}, \boldsymbol{\omega}, \boldsymbol{\psi}) = \mathcal{L}^{\text{recon}}(\boldsymbol{\theta}, \boldsymbol{\omega}) + \mathcal{L}^{\text{inverse}}(\boldsymbol{\theta}, \boldsymbol{\psi})$, where $\mathcal{L}^{\text{recon}}(\boldsymbol{\theta}, \boldsymbol{\omega})$ is the loss function for reconstructing the spatio-temporal concentration field, and $\mathcal{L}^{\text{inverse}}(\boldsymbol{\theta}, \boldsymbol{\psi})$ is the loss function for the inverse problem. The schematic of our proposed multi-task learning framework is shown in Figure \ref{fig:nn_arch}. 




\subsection{Field Reconstruction using Diffusive Masked Convolution}


We employ a series of masked convolutions for the reconstruction of the spatio-temporal concentration field from the sparse observations. We first define a ``masked'' convolution operation by restricting the convolution operation on the masked region centered around the sensor locations where we actually have the observations. The masked convolution operation is defined as,
\begin{align}
\label{eq:partialconv}
    x' = \begin{cases}
      W^{T}(X \odot M) \frac{\text{sum}(\mathds{1})}{\text{sum}(M)} + b, & \text{if sum}(M) > 0 \\
      0, & \text{otherwise}
    \end{cases}
\end{align}
where $W$ and $b$, respectively, denote the weights and bias of a convolutional filter, $X$ is the current feature (pixel) values for the convolutional sliding window, $M$ is a corresponding real-valued mask between 0 and 1 (i.e., $M_{ij} \in [0, 1])$, and
$\odot$ represents element-wise multiplication operation. Note that the masked convolution is defined similar to the Partial Convolutions \cite{liu2018image} introduced for an inpainting problem.

Then, the mask, $M$, is updated using a diffusion process after each masked convolution layer. The diffusion process is modeled by a Spatial Gaussian Convolutional Kernel of size $k \times k$ denoted by $\tilde{W_k} \in \mathbb{R}_{+}^{k \times k}$ as follows:
\begin{align}
    &\tilde{W_k} (i, j) = \exp(-\frac{1}{2\sigma_k^2}\Big[(i-\frac{k-1}{2})^2 + (j-\frac{k-1}{2})^2\Big]); \\ 
    &i, j \in \{0, 1, .., k-1\} \nonumber,
\end{align}
where $i, j$ are the $i$-th row and $j$-th column of the $k \times k$ kernel, $\sigma_k$ is the only learnable parameter and represents the standard deviation of the spatial kernel. In other words, if $\sigma_k$ is large then the spatial-convolution kernel $\tilde{W_k}$ would diffuse the mask values over a larger spatial region. Thus, the learnable-parameter $\sigma_k$ controls the rate of diffusion between the different layers.

The spatial kernel $\tilde{W}_k$ can be repeated across time once (since the size of the kernel along time is one to ensure time-invariance) and the input $C_{in}$ and output $C_{out}$ channels to form the time-invariant Diffusion Kernel $W^{\text{diff}}_k$, and the mask is updated using the diffusion as follows:
\begin{align}
    M' = (W^{\text{diff}}_k)^T M
\end{align}

After the mask update we additionally clip the mask values greater than one. We first initialize the mask at the input layer $M_0$ by the sparse sensor-network locations, as shown in Equation \ref{eq:init_mask}.
\begin{align}
      M_0(\mathbf{x}) = \begin{cases}
      1, & \text{if} \:\: \mathbf{x} \in \mathbf{x^\Phi}\ \\
      0, & \text{otherwise}
    \end{cases}
\label{eq:init_mask}
\end{align}

Then, we stack multiple Diffusive Masked Convolution layers so that the mask $M_0$ gradually grows after each convolution layer and ultimately we can reconstruct the full field from the sparse sensor measurements. Note that this formulation using the Diffusive Masked Convolution can encode arbitrarily placed sensors as the input mask. Thus, by training such a model with different sparse sensor measurements, it is possible to predict the concentration fields on unseen sensor-network configurations.

\subsection{Gaussian Negative Log-likelihood Formulation for Uncertainty Quantification}
In this problem setup, we are also interested in obtaining the uncertainty estimates of the predicted concentration fields. Here, we employ a Gaussian model, where $\phi \sim \mathcal{N}(\phi_{\mu}, \sigma^2)$ to quantify the uncertainty in the predicted concentration field. 
The negative log-likelihood loss function \cite{nix1994estimating} is given as, 
\begin{align}
    \mathcal{L}^{\text{recon}} = \frac{1}{2} \mathbb{E}_{x \sim p_{data}}[\log \sigma^2 + \frac{(\phi_{true}-\phi_{\mu})^2}{\sigma^2}]
\end{align}

where $\phi_{\mu}$ and $\sigma^2$ and denote the predicted mean and variance of the concentration field respectively. Later, in Section \ref{sec:results} we demonstrate that the Gaussian negative log-likelihood loss formulation prevents over-fitting on the ground truth concentration field and offers a smooth estimation of the predicted mean concentration $\phi_{\mu}$.

\subsection{Emission Characteristics Estimation}
Next, we consider the task of solving the inverse problem of estimating the emission characteristics. One of the challenges is that the number of potential sources can vary, thus, the decoder $f_2$ should be able to handle this varying output size. We propose to divide the 2D spatial domain ($N_x \times N_y$) into a $S \times S$ grid. For each cell in the $S \times S$ grid we predict the following source characteristic vector $[p_{i}, x_{i}, y_{i}, c_{i}]$, where $i \in \{1, 2, ..., S^2\}$, $p_{i}$ is the probability of containing a source in the grid cell $i$, $x_{i}$ and $y_{i}$ represent the relative location of the source with respect to the top left corner of the cell, and $c_{i}$ represents the emission strength of the source. 

The objective function for estimating the emission characteristics can be evaluated as follows:
\begin{align}
    \mathcal{L}^{\text{inverse}} = &\lambda_{src} \sum_{i=1}^{S^2} \mathds{1}_{ij}^{src}[(x_i^s-\hat{x}_i)^2+(y_i^s-\hat{y}_i)^2] \nonumber \\ &+ \lambda_{src} \sum_{i=1}^{S^2}  \mathds{1}_{ij}^{src}[(\sigma(\hat{p}_i)-1)^2] \nonumber \\ 
    &+ \lambda_{nosrc} \sum_{i=1}^{S^2}  \mathds{1}_{ij}^{nosrc}[(\sigma(\hat{p}_i))^2] \nonumber \\ &+ \lambda_{src} \sum_{i=1}^{S^2} \mathds{1}_{ij}^{src}[(c_i - \text{Softplus}(\hat{c}_i))^2] 
\end{align}
where, $\mathds{1}_{i}^{src}$ denotes if the source occurs in the grid cell $i$. Note that we use a SoftPlus activation function on the predicted emission strengths $\text{Softplus}(\hat{c}_i)$ to enforce that the emission strengths are always positive, i.e., $c_i \in \mathbb{R}^+$.


\section{Results}
\label{sec:results}

\begin{figure*}[ht]
    \centering
    \includegraphics[width=\textwidth]{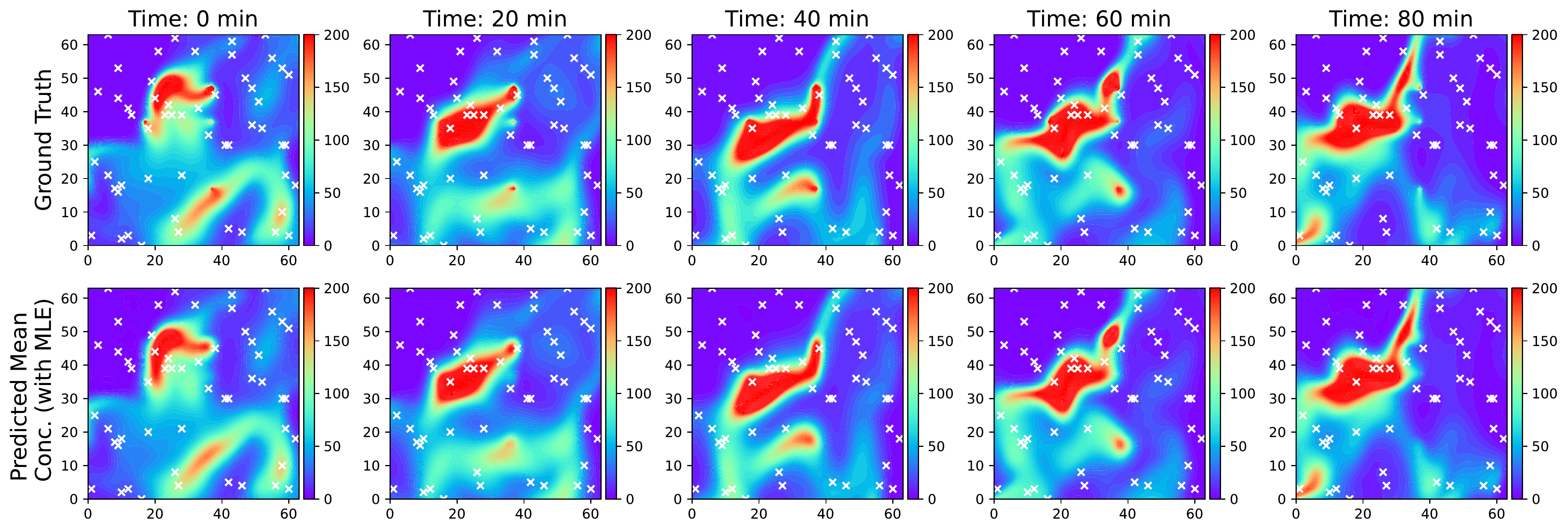}
    \vspace{-3ex}
    \caption{Demonstrating the reconstruction of the global field from sparse sensor measurements on a representative test example. The white crosses denote the position of the sensors. Top row: shows the ground truth concentration fields at various time stamps, Bottom row: the predicted mean of the concentration fields}
    \label{fig:example_recon}
\end{figure*}

\subsection{Experiment and Evaluation Setup} 
\emph{Experimental Setup}: We performed our experiments on 4000 simulations of the forward problem of the advection-diffusion equation, with varying source locations and their emission strengths under stochastically generated wind conditions. In each simulation, the number of sources is randomly selected between one to four, and then randomly placed within the domain. These set of simulations were there divided into training and test sets to train our proposed multi-task learning framework. The proposed multi-task learning framework was trained on $80\%$ of the simulations and tested on the rest. During every epoch, for each simulation in the training set we first randomly select the number of sensors, such that $N_s$ varies between 30 and 100, and then placed them randomly inside the domain. This sampling strategy allows us to generalize on the field reconstruction task from any arbitrary sensor-network during the inference stage.   

\begin{table}[h]
\setlength\tabcolsep{3pt} 
\begin{tabular}{|c|c|c|c|c|c|}
\hline
            & \begin{tabular}[c]{@{}c@{}}Reconstruction \\ RMSE\end{tabular} & Precision & Recall & \begin{tabular}[c]{@{}c@{}}Source Mag. \\ MSE\end{tabular} & \begin{tabular}[c]{@{}c@{}}Rel. Loc. \\ MSE\end{tabular} \\ \hline
w/o G-NLL & 7.96                                                           & 70.6      & 94.2   & 0.178           & 1.38e-7       \\ \hline
w. G-NLL    & 11.29                                                          & 73.3      & 95.6   & 0.209           & 1.32e-7      \\ \hline
\end{tabular}
\caption{Comparing the performance of the model trained with and without the Gaussian Negative Log-likelihood (G-NLL) loss on different evaluation metrics.}
    \label{tab:results_table}
\end{table}

\emph{Evaluation Metrics}: We define the following metrics to evaluate the performance of our model.
\begin{itemize}
    \item \emph{Reconstruction RMSE}: It is defined as the Root Mean-Squared Error (RMSE) between the predicted concentration field and the ground truth concentration field on the test set. A lower value of RMSE indicates higher fidelity in reconstruction of the concentration field.
    \item \emph{Precision and Recall}: To evaluate the reliability of our model in detecting the emission sources, we utilize the precision and recall metrics. For an ideal model, both the precision and recall should be equal to one. Note that the precision and recall are computed on the detections made w.r.t. the $S \times S$ grid.
    \item \emph{Relative Location MSE}: It is defined as the Mean Squared Error (MSE) between the relative locations of the ground truth emission sources and the predicted source locations inside each grid cell. Thus, the Rel. Loc. MSE represents the efficacy of the model in localizing the sources inside the each grid. Therefore, by combining the precision, recall and the Rel. Loc. MSE we are able to fully evaluate the model's capability in localizing the emission sources.
    \item \emph{Source Magnitude MSE}: It is defined as the MSE between the predicted source magnitudes and the ground truth magnitude of the emission sources. It is used as a metric to quantify the emission strength identification ability of the model. 
\end{itemize}


\begin{figure}[h]
    \centering
    \includegraphics[width=0.45\textwidth]{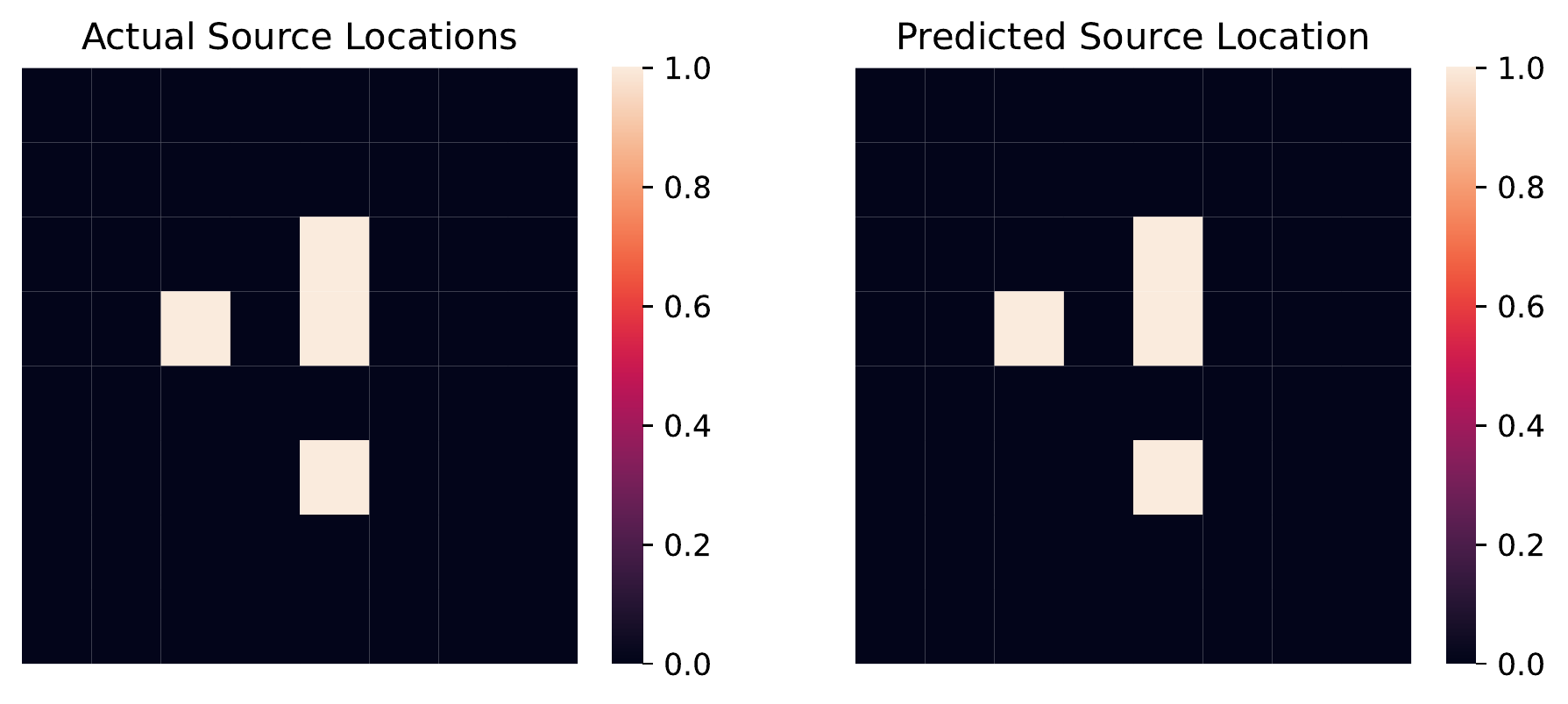}
    \vspace{-1ex}
    \caption{Localization of the pollution sources on a representative test example.}
    \label{fig:example_src_loc}
\end{figure}
\vspace{-2ex}
\begin{figure}[h]
    \centering
    \includegraphics[width=0.26\textwidth]{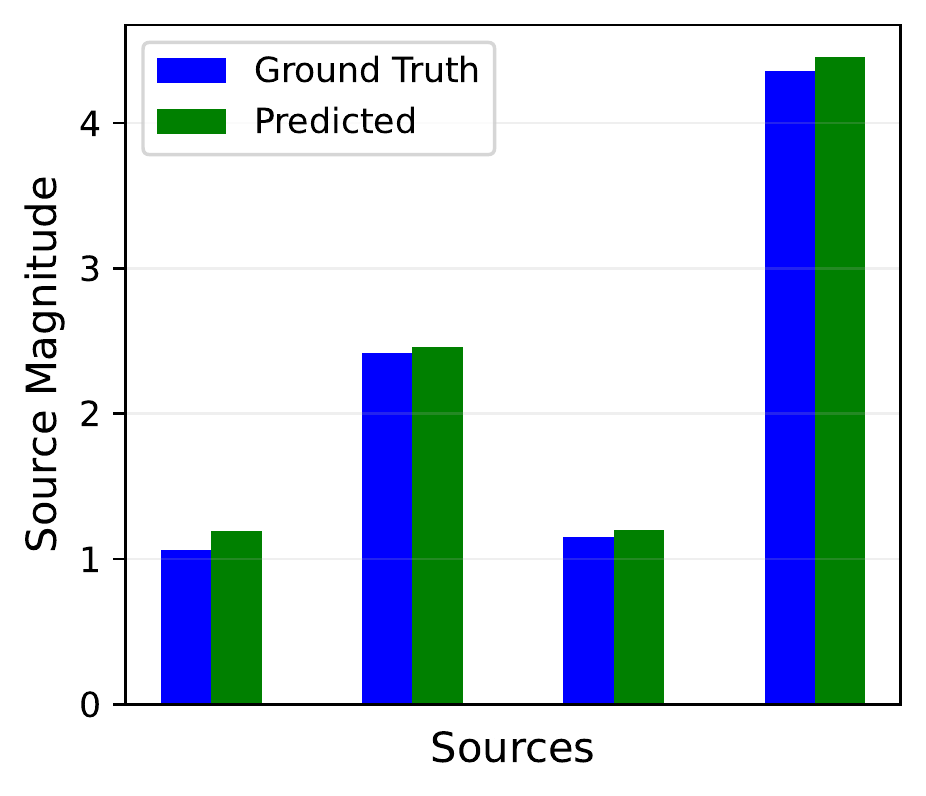}
    \vspace{-1ex}
    \caption{Emission strength estimation on a representative test example.}
    \label{fig:example_src_mag}
\end{figure}

\subsection{Comparing Model Performance}
Table \ref{tab:results_table} shows the comparison of the performance of the model trained with and without the Gaussian Negative Log-Likelihood (G-NLL) Loss on the different evaluation metrics. 

\begin{figure*}[h]
    \centering
    \includegraphics[width=1.0\textwidth]{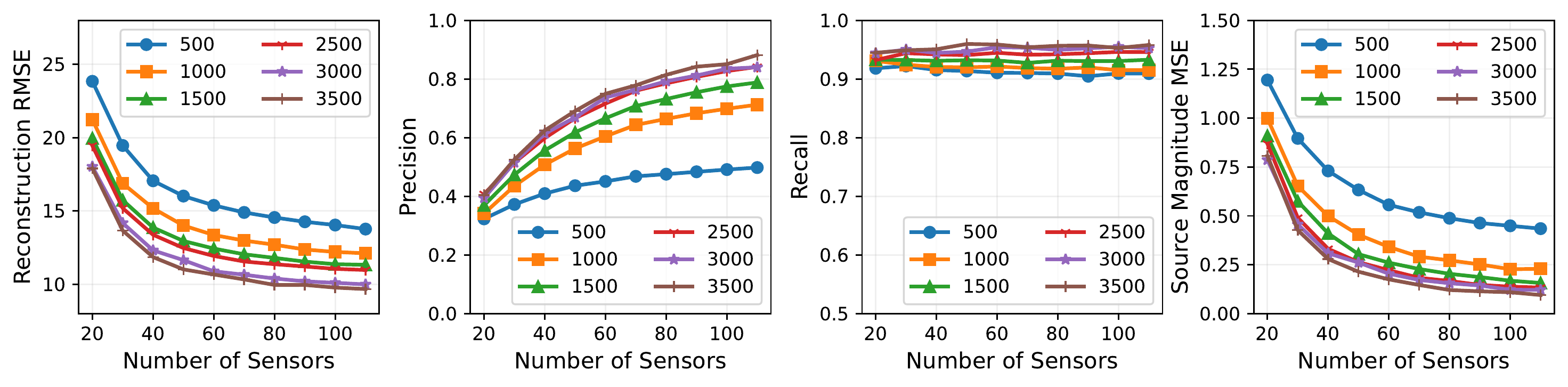}
    \caption{Sensitivity to the number of sensors with varying training set sizes (500-3500 simulations)}
    \label{fig:sensor_sensitivity}
\end{figure*}
\begin{figure*}[h]
    \centering
    \includegraphics[width=1.0\textwidth]{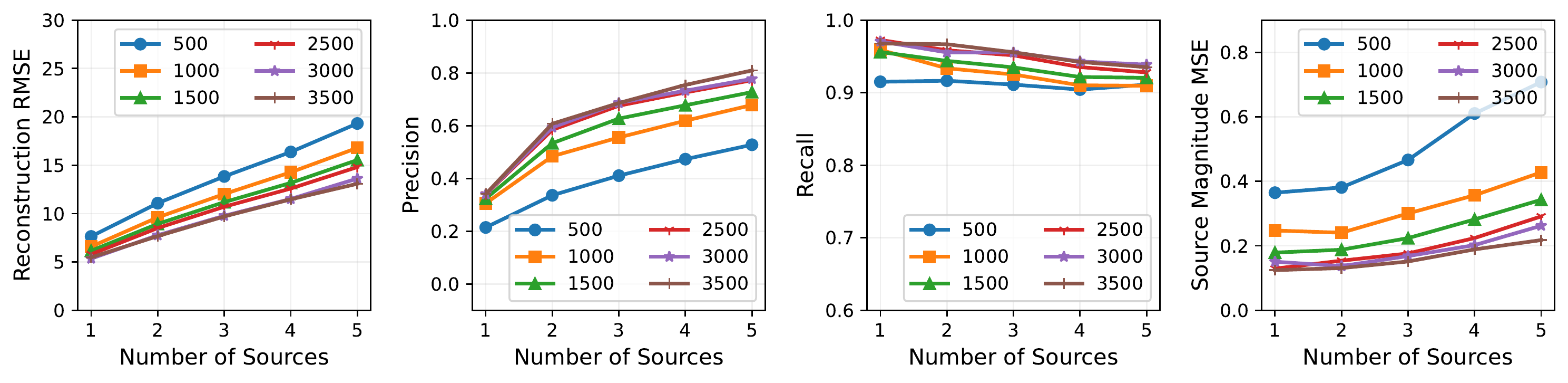}
    \caption{Sensitivity to the number of pollution with varying training set sizes (500-3500 simulations)}
    \label{fig:source_sensitivity}
\end{figure*}

Figure \ref{fig:example_recon} shows an example of the reconstructed concentration field on a test dataset. The proposed diffusive masked convolution models have a Reconstruction RMSE of 7.96 when trained without the Gaussian NLL loss, but the Reconstruction RMSE increases to 11.29 when trained with the Gaussian NLL loss. 
The larger RMSE of G-NLL is mainly due to the misfit at the emission source location. In G-NLL, the estimated concentration field becomes smooth and the delta-function-like behavior at the source location is treated as a noise. On the other hand, without G-NLL, the model tries to follow the delta-function-like behavior, which results in an artificial oscillation in the solution.

We also demonstrate a fairly high recall of around $95\%$, and a precision of $70\%$ for both models. This suggests that the model is able to recover $95\%$ of the emission sources that were present in the simulations. However, the low precision suggests that the model also generates about $30\%$ false positives in its predictions. Further, we observe that once a grid cell detects a emission source, both models can exactly pin-point the location of the source inside the grid with a Rel. Loc. MSE of around $1.3e-7$. 

We also show an accurate estimation of the emission strength for the sources as two models display a Source Mag. MSE of 0.18 and 0.21 respectively. Figure \ref{fig:example_src_loc} and \ref{fig:example_src_mag} shows the ability of our proposed framework to accurately identify the pollution sources as well as approximate their emission strengths on an example case.

\subsection{Analyzing Sensitivity to the Number of Sensors}
Next, we analyze the sensitivity of our model performance on the number of sensors. We vary the number of sensors from 20 to 110 and compare the performance of our model on various training sizes ranging from 500 simulations to 3500 simulations. From Figure \ref{fig:sensor_sensitivity}, it can be seen that all of the evaluation metrics gradually improve as the size of the training set increases. The reconstruction RMSE decreases as the number of sensors increase. This trend is expected as having more sensors observations make the reconstruction problem less ill-posed, thereby making it easier to obtain higher fidelity field reconstructions.  We also observe that the with lower number of sensors the precision decreases significantly while the recall remains the same. This suggests that with less sensors the model is unable to exactly identify the location of the sources and ends up predicting a large number of false-positives that decrease the precision. Additionally, having more sensors improves the Source Mag. MSE, suggesting that the model is able to better estimate the source magnitudes (i.e., solve the inverse problem better) with more sensor measurements.

\begin{figure*}[h]
    \centering
    \includegraphics[width=0.8\textwidth]{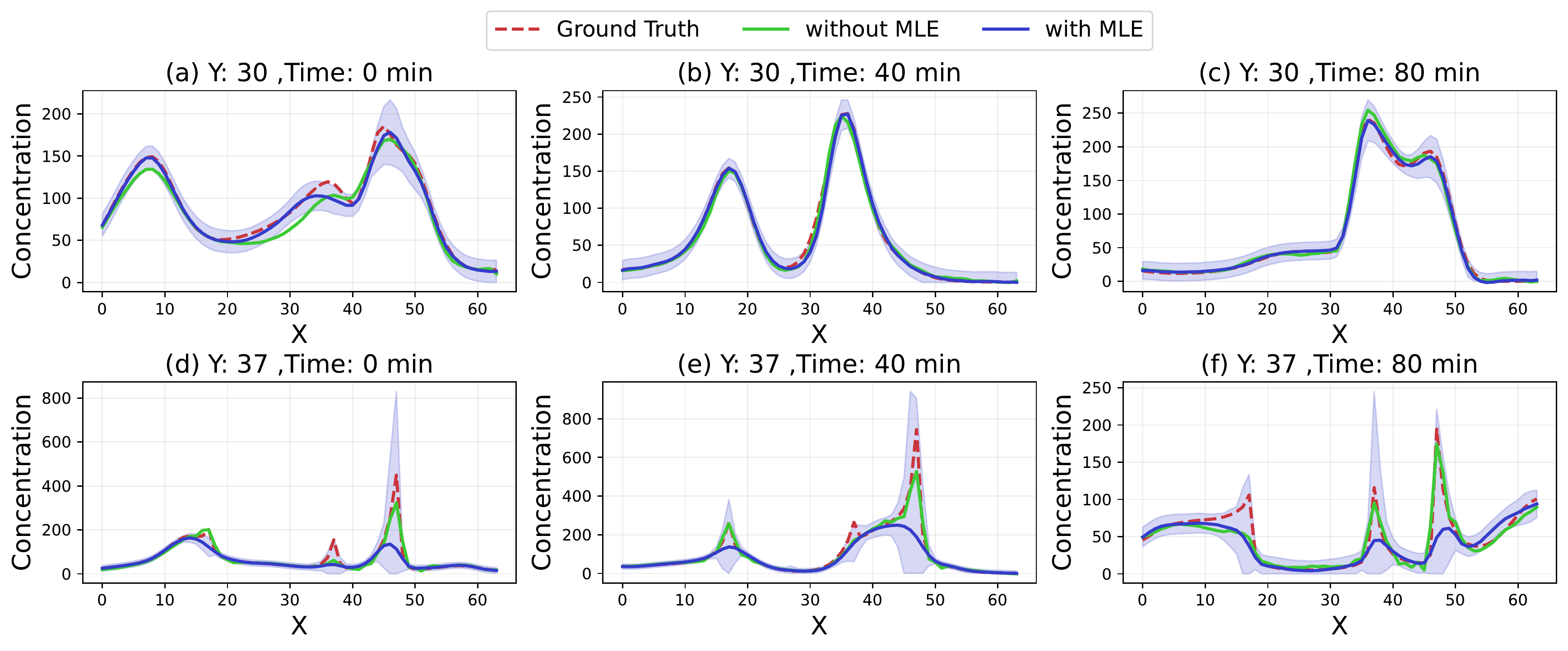}
    \caption{Analyzing the empirical coverage of the 95\% Prediction Interval (shown as the shaded region) for the model trained on Gaussian Negative Log-likelihood (NLL) loss for a particular spatial slice of $y=30$ and $y=37$ (a line passing through 3 pollution sources, that can be identified using the peaks in the ground truth) on an example case.}
    \label{fig:example1_fixed_y}
\end{figure*}

\subsection{Analyzing Sensitivity to the Number of Emission Sources}

Similarly, we analyze the sensitivity of the trained model on the number of emission sources by varying them from one to five sources. From Figure \ref{fig:source_sensitivity}, we can see that the Reconstruction RMSE linearly increases with the number of sources. This is because the magnitude of the concentration field is proportional to the number of emission sources, and thus the same is reflected on the reconstruction error. We also observe that the precision improves as the number of sources increases and the recall remains almost unaffected. Further, the source magnitude MSE increases with the number of sources, as it becomes difficult to disentangle the contributions from multiple emission sources.

\subsection{Effects of the NLL formulation}
\label{sec:analuzing_pred_interval}
\begin{figure}[h]
    \centering
    \includegraphics[width=0.45\textwidth]{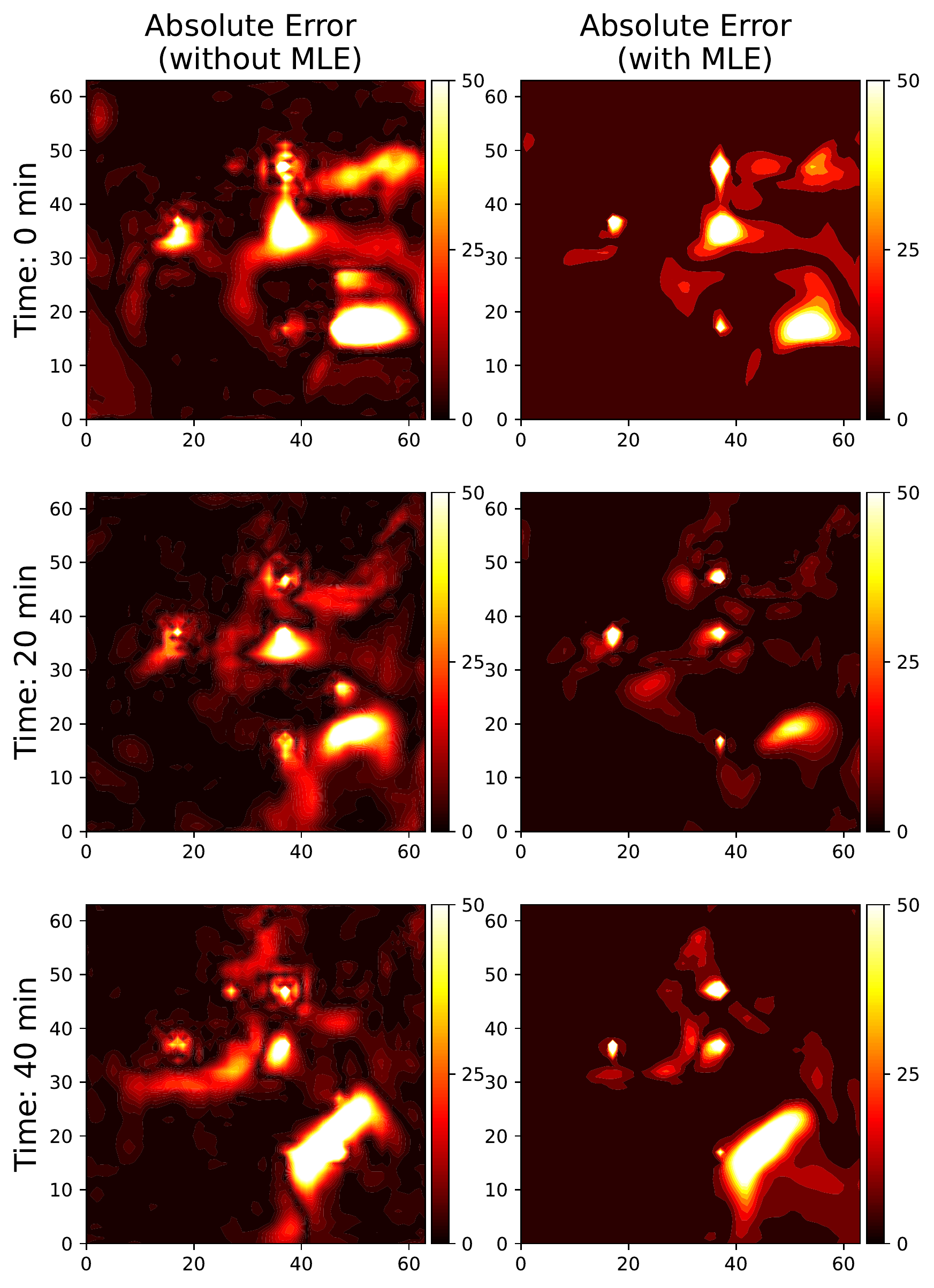}
    \caption{Comparing the Absolute Errors of the model trained with and without the Gaussian Negative Log-likelihood (NLL) loss at different time intervals on an example test case.}
    \label{fig:uq_variance_error}
    \vspace{-2ex}
\end{figure}

We further qualitatively analyze the predictive intervals of the model trained with the Gaussian Negative Log-likelihood loss. For a fixed value of the y-coordinate ($y=30$ and $y=37$) we plot the variations in the concentration fields w.r.t. the x-axis at various times in Figure \ref{fig:example1_fixed_y}. We observe that the predictive intervals estimated by our model always engulf the ground truth concentration. Additionally, we show that the Gaussian-Negative Loglikelihood (NLL) formulation for estimating uncertainties allows the estimate of the mean concentration field to be a smooth function, as the large fluctuations near the pollution sources are taken care by the predicted variance, thereby preventing an oscillatory solution. This behavior can also be seen in Figure \ref{fig:uq_variance_error}, where we see that the absolute error without the NLL shows rectangular oscillatory patterns, which is not observed for the model trained with NLL.

\vspace{-1ex}
\section{Conclusion}

We present a multi-task learning framework for identifying potential emission sources and obtaining reconstruction of spatio-temporal concentration fields from sparse sensor measurements. We also propose a novel diffusive-masked convolution operations that employs the diffusion process in performing masked-convolutions. The diffusive-masked convolution is realized by a spatial-Gaussian convolution kernel, followed by diffusing the mask to nearby regions. Thus, by stacking multiple such layers we are able to iteratively diffuse the information from a sparse sensor measurements (represented using the initial mask) in a principled manner until the learned representations spread over the entire spatio-temporal domain. We also demonstrate precise reconstruction of the concentration field along with accurate localization and emission strength estimation of the pollution sources on the test simulations using our proposed framework. 


\bibliographystyle{plain}
\bibliography{arka}

\end{document}